\documentclass[11pt,a4paper]{article}
\usepackage[]{emnlp2022}
\usepackage{times}
\usepackage{ragged2e}
\usepackage{float}
\usepackage{latexsym}
\usepackage{url}
\usepackage{multirow}
\usepackage[ruled,vlined]{algorithm2e}
\usepackage{algorithmic}
\usepackage[utf8]{inputenc} % allow utf-8 input
\usepackage[T1]{fontenc} % use 8-bit T1 fonts
\usepackage{hyperref} % hyperlinks
\usepackage{url} % simple URL typesetting
\usepackage{amsfonts} % blackboard math symbols
\usepackage{nicefrac} % compact symbols for 1/2, etc.
\usepackage{microtype} % microtypography
\usepackage[utf8]{inputenc} % allow utf-8 input
\usepackage[T1]{fontenc} % use 8-bit T1 fonts
\usepackage{hyperref} % hyperlinks
\usepackage{url} % simple URL typesetting
\usepackage{booktabs} % professional-quality tables
\usepackage{amsfonts} % blackboard math symbols
\usepackage{nicefrac} % compact symbols for 1/2, etc.
\usepackage{microtype} % microtypography
\usepackage{graphicx}
\usepackage{xcolor}
\usepackage[super]{nth}
\usepackage{gensymb}
\usepackage{subcaption}
\usepackage{amsthm}
\usepackage{mathrsfs}
\usepackage{amsmath}
\usepackage{amssymb}
\usepackage{mathtools}
\usepackage{wrapfig}
\usepackage{soul}
\usepackage{pifont}
\usepackage{array}

\makeatletter
\def\BState{\State\hskip-\ALG@thistlm}
\makeatother

\newcommand{\ba}[1]{\begin{align}#1\end{align}}

\newcommand{\minus}{\scalebox{0.5}[1.0]{$-$}}

\usepackage{stackengine}

\makeatletter
\newcommand{\distas}[1]{\mathbin{\overset{#1}{\kern\z@\sim}}}%

\usepackage{algorithm2e}

\newcommand{\beqs}{\vspace{0mm}\begin{eqnarray}}
\newcommand{\eeqs}{\vspace{0mm}\end{eqnarray}}
\newcommand{\barr}{\begin{array}}
\newcommand{\earr}{\end{array}}

\usepackage{bbm}

\usepackage{amssymb}% http://ctan.org/pkg/amssymb
\usepackage{pifont}% http://ctan.org/pkg/pifont

\newcommand{\ours}{{LF}}

\title{
Language Rectified Flow: Advancing Diffusion Language Generation with Probabilistic Flows \\
}

\author{Shujian Zhang  \qquad Lemeng Wu \qquad Chengyue Gong \qquad  Xingchao Liu \\
The University of Texas at Austin\\
\texttt{\{szhang19, lm.wu, cygong, xcliu\}@utexas.edu}
}

\begin{document}
\maketitle

\begin{abstract} 
Recent works have demonstrated success in controlling sentence attributes ($e.g.$, sentiment) and structure ($e.g.$, syntactic structure) based on the diffusion language model. A key component that drives the
impressive performance for generating high-quality samples from noise is iteratively denoise for thousands of steps. While beneficial, the complexity of starting from the noise and the learning steps has limited its implementation to many NLP real-world applications. This paper proposes Language Rectified Flow ({\ours}).
Our method is based on the reformulation of the standard probabilistic flow models.
Language rectified flow learns (neural) ordinary differential
equation models to transport between the source distribution and the target distribution, hence
providing a unified and effective solution to generative modeling and domain transfer.
From the source distribution, our language rectified flow yields fast simulation and effectively decreases the inference time. 
Experiments on three challenging fine-grained control tasks and multiple high-quality text editing show that our method consistently outperforms its baselines. Extensive experiments and ablation studies demonstrate that our method can be general, effective, and beneficial for many NLP tasks. 
\end{abstract}

\section{Introduction}\label{sec:introduction}
Traditional pretrained large-scale language models (LM) can generate high-quality text 
% by  the pretrained language model 
for specific real-world applications \cite{radford2019language,brown2020language,chowdhery2022palm, zhang2022opt, zhang2023automl}.
However, updating the LM parameters or finding proper prompts for each control task can be expensive and unscalable given the combinatorially many possible compositions and the lack of supervised data. 

Recent research thus has started to explore plug-and-play solutions. With a pretrained language model (LM), the plug-and-play approaches \cite{ krause2020gedi, kumar2021controlled, yang2021fudge, zhang2022passage, mireshghallah2022mix} are served as the light-weight constrained guidance to the targeted text sequence generations \cite{dathathri2019plug}. The approaches, however,
typically rely on search or optimization in the complex text sequence space. The discrete nature of
text makes the search/optimization extremely difficult. Though some recent work introduces continuous approximations to the discrete tokens \cite{kumar2021controlled,qin2022cold}, the high dimensionality and complexity of the sequence space still render it inefficient to find accurate high quality text. The most recent approach Diffusion LM \cite{li2022diffusion} induces continuous latent representations,
which enables efficient gradient-based methods for the controllable generation.

However, despite the advantage of Diffusion LM for separating the distribution map learning from a noise distribution to a meaningful shape distribution, it always requires thousands of steps. The transport trajectory learns from a noise distribution to a meaningful shape
the distribution. This is a major efficiency bottleneck during inference since a standard diffusion model requires thousands of generation steps and a proper SDE solver to produce high-reconstruction and diverse text.
In addition, the existing denoising diffusion techniques requires substantial hyper-parameter search in involved design space.
% and are still poorly understood both empirically and theoretically.

We propose language rectified flow, a surprisingly simple approach to the transport mapping problem, which unifiedly solves both generative modeling and domain transfer. The language rectified flow is an ordinary differential
equation (ODE) model that transports distribution source distribution $\pi_0$ to target distribution $\pi_1$ by following straight line paths as much as possible. The straight paths are
theoretically preferred because it is the shortest path between two end points, and computationally preferred because it can be precisely simulated without time discretization. Hence, flows with straight paths bridge the
gap between few-step and continuous-time models. Specifically, we  formulate an ODE transport flow as the initial text generator with a simpler trajectory
compared with the diffusion model formulated in SDE. Meanwhile, 
we optimize the transport flow cost for the initial flow model
to significantly straighten the learning trajectory while maintaining the model’s performance. Further, a VAE is utilized to connect the text sequence space and the flow latent space. This can be used to produce the initial values in the language flow for our controllable text generation.

To demonstrate control of language rectified flow, we consider three control targets ranging from span-anchored
controls (e.g., length control and infilling) to complex structures (e.g., parse trees). Our {\ours} can generate high-quality text, performing favorably relative to the diffusion-based LM with a 27x
faster sampling on the controllable text generation
task.  In addition to these individual control tasks, we conduct experiments on three challenging
settings, including sequential editing of text. Results show that composing operators within our method manage to generate or
edit high-quality text, substantially improving over
respective baselines in terms of quality and efficiency. Furthermore, we provide extensive ablation studies on different design choices for the proposed method, including the evidence with different generation steps, influence of the constrained optimization, flow in the latent space. Our analysis shows that {\ours} contributes the performance improvement, helping the sampling efficiency and generation. With little modification, ours can be easily
applied to other NLP tasks for better controllable text generation. 
Our contributions are summarized as follows:
\begin{itemize}
  \setlength{\itemsep}{0pt}
\setlength{\parsep}{0pt}
\setlength{\parskip}{0pt}
\setlength{\parskip}{0pt}
    \item Present a language rectified flow for controllable text generation embracing
    domain transfer and fast simulation with the ordinary differential equation. 
    \item Propose an efficient and effective way to train the language rectified flow, which can optimize the trade-off between representation learning and flow matching.
    % We consider the constrained optimization with a special emphasis on lexicographic optimization. 
    \item Verify the effectiveness and general applicability of the proposed method in various NLP tasks, \emph{$e.g.$}, fine-grained text generation and text editing benchmarks, and provide a rich analysis of our method with various design choices.
\end{itemize}

\section{Method} \label{sec:method_section}
We now introduce our method, Language Rectified Flow ({\ours}),
the transport flow for the controlled language generation. Generating text with transport flow can be viewed as
transporting from source distributions to target distributions
by following a learned trajectory.  
Specifically, we suggest a general recipe for the language rectified flow: 1) construct the continuous latent space, 2) learn a neural velocity flow network, construct an ODE process in the latent space with the shortest transport path, and utilize a neural network to fit this process, 3) propose the lexicographic (lexico) optimization strategy for the joint learning. 
In this section, we present in detail our proposed
method, {\ours} (Algorithm \ref{alg:acquisition}).

\subsection{Encoding and Latent Space}
As the text input is discrete, encoding the language as the latent vector serves as
the higher-level and differentiable sentence representations \cite{dai2019style, li2020optimus}.
Variational auto-encoders (VAEs) \cite{kingma2013auto, mai2020plug}
have been used to model text with a low-dimensional continuous latent space with certain regularities \cite{bowman2015generating}. A VAE connects the text sequence space $\mathcal{X}$ and the latent space with an encoder $q(z|x)$ that maps text $x$ into latent vector $z$, and a decoder $p(x|z)$ that maps a $z$
into text. We learn text VAEs from scratch, optimizing the encoder and decoder parameters with the following objective:
\begin{align}
\begin{split}
& \mathcal{L}_{\mathrm{VAE}}(\boldsymbol{h}) = -\mathbb{E}_{q(\boldsymbol{z} \mid \boldsymbol{x})}[\log p(\boldsymbol{x} \mid \boldsymbol{z})] \\ 
& \quad \quad \quad \quad +\mathrm{KL}\left(q(\boldsymbol{z} \mid \boldsymbol{x}) \| p_{\text {prior }}(\boldsymbol{z})\right)
,
\label{equ:vae_kl}
\end{split}
\end{align}
where $p_{prior}(z)$ is a standard Gaussian distribution
as the prior, and $\mathrm{KL}(\cdot \| \cdot)$ is the Kullback-Leibler divergence that pushes $q_{enc}$ to be close to the prior. The first term encourages $z$ to encode relevant information for reconstructing the observed text $x$,
while the second term adds regularity so that any
$z \sim p_{prior}(z)$ can be decoded into high-quality text in the text sequence space $\mathcal{Z}$. Finally, VAE decoder $p(x | z)$ offers a way to map any given latent vector z into
the corresponding text sequence.

\begin{figure*}[t]
\centering
\includegraphics[width=15.0cm]{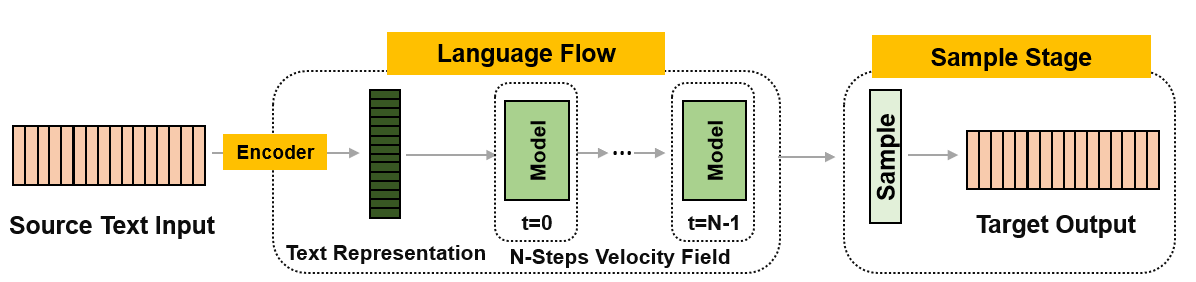}
\caption{Overview of language rectified flow. Some notations are labeled along with corresponding components. 
% `Passage Rep' refers to the passage representation, `Mask' refers to the mask, `e' refers to the encoder, `d' refers to the decoder, and `Q + P$_1$' refers to the question and the first passage. In the Mask, the black color represents the mask and the white color represents the non-mask.
}
\label{fig:pipeline}
\end{figure*}

\subsection{Probability Flows}
% \sz{flow 
% background forumulation 
% 1。 diffusion lanuage flow
% 2. bridge iclr 2022 Language modeling via stochastic processes
% }

% \sz{alogrithm 里面加上对应的例子}

% \sz{在这两段里面需要说一下上面的背景，然后稍微说一下 diffusion 和bridge 是sde 来源于noise 所以更容易generalize，但是ode 不是来源于noise 但是更容易train，所以效果训练的快，然后效果也好}

% \sz{diffusion， langugage， 有的更容易训 没有

% noise 可能更容易generalize，
% }

A stochastic differential equation (SDE) characterizes a diffusion process that maps real data to random noise in continuous
time $t \in [0, T]$. Specifically, let $z_t$ be the variable of the process following $z_t \sim \pi_t$, indexed by time $t$. At start time $t = 0$, $z_0 \sim \pi_0$ which is the data distribution, and at the end $t = T$, $z_T \sim \pi_T$ which is the noise distribution ($e.g.$, standard Gaussian). The reverse SDE instead generates a real sample from the noise by working backwards in time (from $t = T$ to $t = 0$).  This diffusion process can be modeled as the SDE: 
\ba{
\mathrm{d} z_t=v_\theta\left(z_t, t\right) \mathrm{d} t + g_t dw, \quad z_0 \sim \pi_0
,
\label{eq:sde}
}
where $g_t$ is a scalar function known as the diffusion coefficient of $z_t$ and SDE adds a stochastic term $w$ (the standard Wiener process (a.k.a., Brownian motion)) in each update.
% which makes it hard to simplify.
The recent successful diffusion models \cite{song2020score, ho2020denoising} can be understood as learning stochastic differential equations (SDEs) \cite{anderson1982reverse}, and lots of SDE techniques have been applied for developing better models \cite{karras2022elucidating}. 
In practice, an SDE transport flow usually needs thousands of steps to reach the target distribution. In this work, instead of focusing on SDE, language rectified flow aims to build ODE with a simple training objective and fast sampling.

Our goal is to build a transport flow to push the latent text input from the source distribution to the target distribution. 
Specifically, given \emph{i.i.d.} samples $\mathcal{D}=\left\{z^{(i)}\right\}_{i=1}^N$, we denote $z_0$ as the data samples from the source distribution $\pi_0$ and $z_1$ as the data samples from the complex
target distribution $\pi_1$. A probability flow can be effectively learned by training a velocity field $v_\theta$ of an ordinary differential equation (ODE), which is indexed by a continuous time variable $t \in [0, T]$ as below:
\ba{
\mathrm{d} z_t=v_\theta\left(z_t, t\right) \mathrm{d} t, \quad z_0 \sim \pi_0
,
\label{eq:initial}
}
where $z_t$ is the intermediate state representation at time $t$ and it serves as the linear interpolation of $z_0$ and $z_1$. The velocity field $v_\theta(z_t, t)$ is a neural network parameterized by $\theta$. The drift force $v: \mathbb{R}^d \rightarrow \mathbb{R}^d$ is set to drive the flow to follow the direction $(z_1 - z_0)$ of the linear path pointing from $z_0$ to $z_1$ at any given time $t$ as much as possible. To optimize the velocity, with $t\in[0, 1]$,  the flow is followed the ODE process $\mathrm{d}x_t = (z_1-z_0)\mathrm{d}t$ by solving a least squares regression problem as $\min_\theta \int_0^1 \mathbb{E}_{z \sim \mathcal{D}} \bigg [\|(v_\theta(z_t, t) - (z_1 - z_0)\|^2 \bigg]\mathrm{d}t$
$\mathrm{with}~~ z_t = t z_1 + (1-t)z_0 $.
During the training, we discretize the above as
\begin{align}
\resizebox{0.95\hsize}{!}{$\mathcal{L}_{\mathrm{FLOW}}=
\min_\theta \mathbb{E}_{z \sim \mathcal{D}, t \sim [0, 1]} \bigg[\|(v_\theta(z_t, t) - (z_1 - z_0)\|^2   \bigg].$}
\label{eq:ode}
\end{align}
By fitting the drift $v_\theta$ with $z_1 - z_0$, the language flow causalizes the paths of linear interpolation $z_t$, yielding an ODE flow that can be simulated.
After we get $v$, we solve the ODE starting from $z_0 \sim \pi_0$ to transfer $\pi_0$ to $\pi_1$, backwardly starting from $z_1 \sim \pi_1$ to transfer $\pi_1$ to $\pi_0$. The forward and backward sampling are
equally favored by the training algorithm, because the objective in Eqn \eqref{eq:ode} is time-symmetric in that it yields the
equivalent problem if we exchange $z_0$ and $z_1$ and flip the sign of $v$.

\subsection{Efficiency with Language Rectified Flows}\label{sec:eff_language_flow}
After the neural velocity field $v_\theta$ is well-trained, samples
can be generated from the ODE to get $z_1$ as draws from the target distribution $\pi_1$ by discretizing the ODE process with Euler solver in Eqn~\eqref{eq:initial} into N steps,
\begin{equation}
% \label{eq:euler}
    z_{(k+1)/N} \leftarrow z_{k/N} + \frac{1}{N} v_\theta(z_{k/N}, k/N),
    \label{eq:sample}
\end{equation}
%where we use a grid of time points $\left\{t_k = k/N \right \}_{k=0}^{N}$. $x_0=x_{t_0}$ is a random sample from $\pi_0$ and $x_1=x_{t_N}$ is the generated data.
where the integer time step $\hat{t}$ is defined as $\hat{t} \in \{0,1,\cdots,N-1\}$. 
$k$ is defined as $k \in \{0, 1,\dots, N-1\}$. Here $z_0 = z_{0/N}$ is a random sample from $\pi_0$ and $z_1 = z_{N/N}$ is the generated data.
The number of discretization steps, $N$, determines the closeness of the simulated trajectory Eqn~\eqref{eq:sample} and the learned continuous ODE trajectory. When $N \rightarrow \infty$, the simulated trajectory Eqn~\eqref{eq:sample} has approached the same endpoint as the continuous one. Intuitively, the solver will be more accurate with a larger $N$.
The language ODE flow follows straight line paths as much as possible. Compared to SDE ($e.g.$, diffusion language model usually requires 1000 to 2000 steps), the straight paths are
preferred both theoretically because it is the shortest path between two end points \cite{lipman2022flow, liu2022flow, liu2023flowgrad, liu2023instaflow}, and computationally
because it can be exactly simulated without time discretization.
% It unifiedly solves both generative modeling and domain transfer. 
In practice, it is often found that an appropriate choice of $N$ for existing probability flow ODEs ranges from $10$ to $20$. 

\subsection{Constrained Optimization}
\paragraph{Trade-off is a Problem.} 
The encoder-decoder network aims to encode and decode the representation of input text and the flow network aims to transfer from the source distribution to the target distribution. 
We consider this problem with a bi-level optimization perspective: we want to identify the ideal flow path within the optimized latent representation of the input text.
We solve the problem with a simple gradient descent-like approach that iteratively updates the flow network and the encoder-decoder network
in a guaranteed fashion. The key feature of our method
is that it ensures to optimize the reconstruction loss as a typical
optimization method while adding transfer from the source distribution to the target distribution as an extra constrained loss, 
\ba{
    \min \mathcal{L}_{\mathrm{VAE}} ~~\mathrm{s.t.}~~ \mathcal{L}_\mathrm{FLOW} < \mathrm{Constraint},
   % \mathcal{L} = \mathcal{L}_{\mathrm{VAE}} + \lambda \mathcal{L}_{\mathrm{FLOW}}, 
   \label{eq:our_reward}
}
where $\mathcal{L}_{\mathrm{VAE}}$ refers to the reconstruction loss,  $\mathcal{L}_{\mathrm{FLOW}}$ is referred to the Eqn~\eqref{eq:ode}, and $\mathrm{Constraint}$ is a constraint threshold.
The linear combination of multiple objectives is the most widely used approach. 
However, the coefficient of the combination requires manual tuning, and it is  theoretically unsuitable for non-convex functions. This work considers constrained optimization on trading off two objectives, with a special emphasis on lexicographic (lexico) optimization.

\paragraph{Our Equation.}
To optimize the trade-off between flow optimization and representation construction in Eqn \eqref{eq:our_reward}, 
% we propose to use lexicographic optimization, in which the parameters are iteratively updated as 
% \ba{
% \theta_{t+1} \leftarrow \theta_t-\gamma_t e_t,
% \label{eq:our_pareto}
% }
% where $\gamma_t \geq  0$ is an adaptive step size and $e_t \in \mathbb{R}^{d}$ is an update direction to be chosen to balance the minimization of $f$ and constraint satisfaction on $q$.
% One of the objectives (say $f$ which is $\mathrm{flow}$ in our case) is of secondary
% importance w.r.t. the other one (say $q$  which is $\mathrm{reconstruction}$).
% The design criterion for the constrained optimization is when the constraint is not satisfied (i.e., $q(\theta_t) \ge c$), the focus becomes decreasing $q$ to satisfy the constraint as soon as possible; in the meantime, $f$ performs as a secondary objective indicating that $f$ should be minimized to the degree that it does not hurt the descent of $q$.
% Therefore, 
we use lexicographic optimization, in which the parameters are iteratively updated to obtain such a goal: 
\ba{
\theta_{s+1} \leftarrow \theta_s-&\gamma_s (\nabla \mathcal{L}_{\mathrm{VAE}}  %\times
\notag\\
&%\left(\nabla \mathrm{quality} \left(\theta_t\right)
~~~~+\lambda_s \nabla \mathcal{L}_{\mathrm{FLOW}}  \left(\theta_s\right)), %\right),
\label{eq:our_pareto_2}
}
where $\gamma_s \geq  0$ is an adaptive step size, $\nabla \mathcal{L}_{\mathrm{FLOW}}$ and $\nabla \mathcal{L}_{\mathrm{VAE}}$ are estimated by score function, and the $\lambda_s$ can be computed as 
$
\lambda_s=\max \left(\frac{\phi\left(\theta_t\right)-\nabla \mathcal{L}_{\mathrm{VAE}}\left(\theta_s\right)^{\top} \nabla \mathcal{L}_{\mathrm{FLOW}} \left(\theta_s\right)}{\left\|\nabla \mathcal{L}_{\mathrm{FLOW}} \left(\theta_s\right)\right\|^2}, \quad 0\right).$ 
$\phi (\theta_s)$ equals to $q(\theta_s)-c$ and the $c$ represents the minimal loss and $s$ is the number of optimization iterations.

\paragraph{The Proposed Algorithm.} Our text flow with sampling efficiency and lexico optimization is 
shown in Algorithm \ref{alg:acquisition}.
We iteratively update the language rectified flow and the latent representation network in a single-loop manner.
% The policy network parameter $\theta$ is updated by Eqn~\eqref{eq:our_pareto} in a direction to balance the optimization of quality and constraint satisfaction on computation.
Overall, our algorithm gives a ten-step text generation approach. 
After these three training stages, one can sample a language rectified flow in a few steps starting from a source domain by following Eqn~\eqref{eq:sample}.
% Random sample an initial poi

\begin{algorithm*}[t]
%  \scalebox{0.7}{
\caption{\small Language Rectified Flow} 
\label{alg:acquisition}
\begin{algorithmic}[1]
\footnotesize 
\STATE \textbf{Input:} Source text $x_0 \in \mathcal{X}_0$ and target text $x_1 \in \mathcal{X}_1$. The initial velocity field $v_\theta$ parameterized by $\theta$.
%for input $\bx$.
\STATE \textbf{Training Stage}
\FOR{s iterations}
\STATE Randomly sample  $x_0 \sim \mathcal{X}_0$  and  $x_1 \sim \mathcal{X}_1$.
\STATE Encode the $x_0, x_1$ as latent vector $z_0 \in \pi_0$, $z_1 \in \pi_1$.
% Generate $K$ augmentations,  $\left\{\bx_{i}^{\prime}\right\}_{i=1, \cdots, K} \leftarrow g\left(\bx\right)$. \aux{data paraphrasing via augmentation}
\STATE % Construct the probability ODE flow as Eqn \eqref{eq:initial} and 
Train  $v_\theta$ follows the objective function Eqn~\eqref{eq:ode} with $t$ uniformly sampling from $[0, 1]$.
% Compute $p_\theta(\cdot \mid \bx)$ and  $p_\theta(\cdot \mid \bx^{\prime}_{i})$ for $i=1, \ldots, K$. \aux{compute probabilities}
\STATE
Update $\theta$ with the lexicographic optimization in Eqn \eqref{eq:our_pareto_2}.
% \STATE
%  \sz{Randomly sample from $X'_0 \sim \mathcal{N}(\textbf{0}, \textbf{I})$, and output the desired point cloud $X'_1$ with $X'_1 = X'_0 + v_\theta(X'_0, 0)$.}
 \ENDFOR
\STATE \textbf{Sample Stage}
\STATE Given a $x_0 \in \mathcal{X}_0$, encode it to $z_0 \in \pi_0$, transfer it to domain $\pi_1$ using Eqn~\eqref{eq:sample} and well-trained $v_\theta$. Then, decode it to target text domain $\mathcal{X}_1$.
%  \STATE
% Select top $s_\textit{acq}$ largest examples in $\cD_{pool}$, according to the value of $\mathbb{D}(p_\theta(\cdot \mid \boldsymbol{x}), p_\theta(\cdot \mid \boldsymbol{x^{\prime}}))$.
%  \STATE Label these $s_\textit{acq}$ examples.
%  \STATE \textbf{end while}
%  %of their generated images.
% %, save them as $\{\vec z_i, \vec y_i\}_{i=1}^k$.
% \STATE Curriculum learning the model parameters with  
% Eqn \eqref{eq:ssl}.
\end{algorithmic}
\end{algorithm*}

\section{Experimental Settings}\label{sec:experiemental_settings}
We evaluate our method on control tasks and text editing tasks. Table \ref{tab:dataset_setting} shows the  experimental data configuration. 
\subsection{Control Tasks and Evaluation Metrics}\label{sec:control_experiemental_settings}
\paragraph{Dataset.} We consider three control tasks shown in Table \ref{tab:dataset_setting}: the first two tasks (parts-of-speech and length) rely on the E2E dataset \cite{novikova2017e2e}, and the last task (infill) is based on Abductive NLG dataset \cite{bhagavatula2019abductive}.
For E2E, the dataset provides information about restaurants and consists of more than
50K combinations of dialogue-act-based meaning representations.
% The dataset is split into training, validation, and testing sets (in a 76.5-8.5-15 ratio), keeping a similar distribution of MR and reference text lengths.
% and ensuring that meaning representations in different sets are distinct. 
Abductive NLG (aNLG) is a conditional generation task for explaining given observations in natural language. It is based on the ART \cite{bhagavatula2019abductive} that consists of over 20k
commonsense narrative contexts and 200k explanations. 
% The train set includes all plausible and implausible hypotheses collected via crowdsourcing, while the dev and test sets include the hypotheses
% selected through the Adversarial Filtering algorithm. 

\paragraph{Setting and Metrics.} Following \citet{li2022diffusion}, for each control task, we sample 200 control targets from the validation splits, and we generate 50 samples for each control target. To evaluate the fluency of the generated text, following the prior works \cite{dathathri2019plug, yang2021fudge, li2022diffusion},
we report the perplexity (PPL) of generated
text. In prior works \cite{li2022diffusion}, this metric is named as fluency score. A lower perplexity score indicates better sample quality. 

We define success metrics (SR) for each control task as follows:
\ding{182} For Parts-of-speech, given a sequence of parts-of-speech (POS) tags (e.g., Pronoun Verb Determiner
Noun), generate a sequence of words of the same length whose POS tags (under an oracle POS tagger) match the target (e.g., I ate an apple). We quantify success via word-level exact match. 
\ding{183} For length, given a target length 10, $\cdots$ ,40, generate a sequence with a length within $\pm$2 of the target. 
\ding{184} For infilling, given a left context ($O_1$) and a right context ($O_2$) from the aNLG dataset, and the goal
is to generate a sentence that logically connects $O_1$ and $O_2$. For evaluation, we report automatic metrics (BLEU \cite{papineni2002bleu}, ROUGE \cite{lin2003automatic}, and BertScore \cite{zhang2019bertscore}).

\paragraph{Baselines.}
For POS and length, we compare our method with FT \cite{radford2019language}, FUDGE \cite{yang2021fudge}, and Diffusion LM (DLM) \cite{li2022diffusion}.
FT is a fine-tune GPT-2 on text pairs, yielding an oracle conditional language model \cite{li2022diffusion}.
FUDGE is controllable generation approach based on an
autoregressive LM. Diffusion LM is a diffusion based language model. 
For infilling, three baseline methods are compared including Delorean \cite{qin2020back}, Cold \cite{qin2022cold}, and Diffusion LM \cite{li2022diffusion}.

\subsection{Text Editing Task and Evaluation Metrics}
\paragraph{Dataset.} We evaluate in two domains, including the
Yelp review \cite{shen2017style} preprocessed by \citet{li2018delete} and the Amazon comment corpus \cite{he2016ups}.
For Yelp, each example is a sentence from a business review on Yelp, and is labeled as having either positive or negative sentiment.
Amazon dataset is similar to Yelp. Each example is a sentence from a product review on Amazon, and is labeled as having either positive or negative sentiment \cite{he2016ups}.

\begin{table}[h]
\centering
\footnotesize
\resizebox{1.0\columnwidth}{!}{
 \begin{tabular}{l|c|c|c|c}
 \toprule
 \bf{Task} &\bf{Dataset} &   \bf{Train} & \bf{Val}  & \bf{Test} \\ \midrule
   \multirow{2}{*}{Control Task} & E2E & 42.1K & 4.7K & 4.7K \\
 & Abductive NLG & 256.6K & 4.6K & 9.2K \\
 \hline 
   \multirow{2}{*}{Text Editing} & Yelp & 450K & 4K & 1K \\
 & Amazon & 555K & 2K & 1K \\
%  \hline 
%  Dialogue & Wizard of Wikipedia  & 63.7K & 3.1K & 2.9K \\
% \hline 
%  Fact Verification & FaVIQ-Ambig (A) & 17.0K & 4.3K & 4.7K \\
 \bottomrule
  \end{tabular}}
 \caption{Dataset Configuration. The top block is for the control task and the bottom block is for the text editing.} 
    \label{tab:dataset_setting}
    \vspace{-15pt}
\end{table}

\paragraph{Setting and Metrics.} 
We conduct the sequential editing whose goal is to edit the given text by changing
an attribute each time and keep the main content
consistent. For example, we consider the input sentence as the source distribution with negative sentiment and the goal is to transfer the input sentence from the negative sentiment to positive sentiment.
Generation accuracy is given by a BERT
classifier to evaluate the success rate \cite{zhang2019bertscore}. Perplexity
(PPL) is calculated on the
corresponding domain to measure fluency. 
% ROUGE \cite{lin2003automatic} is also adopted.
% We report the unigram ROUGE1 (R-1) and bigram ROUGE-2 (R-2) overlap to assess the informativeness, and the longest common subsequence ROUGE-L (R-L) score to assess the fluency. 
To further evaluate the ability of content
preservation, we measure BLEU \cite{papineni2002bleu} between human-annotated ground truth and output. For each case, we sample 100 examples to evaluate.

\begin{table*}[tbhp]
\centering
\scalebox{0.85}{
 \begin{subtable}[t]{0.5\linewidth}
 \centering
 \raggedright
 \setlength{\extrarowheight}{6pt}
 \setlength{\tabcolsep}{9pt}
 \begin{tabular}{l|cccc}
 \toprule
\multirow{2}{*}{Model}  & \multicolumn{2}{c}{Parts of Speech} & \multicolumn{2}{c}{Length}\\ 
  \cline{2-5}
 & SR $\uparrow$ &  PPL $\downarrow$ & SR $\uparrow$  &  PPL $\downarrow$\\
   \hline
 FUDGE 
 & 27.0 & 7.96 & 46.9 & 3.11 \\
 FT  
 & 89.5 & 4.72 & 98.1 & 3.84 \\
 DLM 
 & 90.0 & 5.16  & {\bf 99.9} & 2.16  \\
\hline
  {\bf Ours} & {\bf 94.2} & {\bf 4.65} & { 99.3} &{\bf 1.74} \\
  \bottomrule
  \end{tabular}
  \end{subtable}
  \begin{subtable}[t]{0.5\linewidth}
    \centering
    \raggedleft
    \setlength{\extrarowheight}{6pt}
    \setlength{\tabcolsep}{5pt}
  \begin{tabular}{l|cccc}
   \toprule
  \multicolumn{2}{c|}{~} & \multicolumn{3}{|c}{Infill} \\ 
  \cline{3-5}
  \multicolumn{2}{c|}{~} & BLEU $\uparrow$ &  ROUGE-L $\uparrow$ & BERTScore $\uparrow$\\
   \hline 
    \multicolumn{2}{c|}{DELOREAN}
 & 1.6 & 19.1 & 41.7 \\
 \multicolumn{2}{c|}{COLD} 
 & 1.8 & 19.5 & 42.7 \\
 \multicolumn{2}{c|}{DLM} 
 & 7.1 & 28.3  & { 89.0 } \\
 \hline
 \multicolumn{2}{c|}{\bf Ours} & {\bf 8.2} & {\bf 32.7} & {\bf 92.1}\\
 \bottomrule
  \end{tabular}
  \end{subtable}
  }
 \caption{Comparison to models on Parts of Speech, Length, and Infill. `SR' and ` PPL' refer to the success metric and the fluency of the generated text, in Section \ref{sec:control_experiemental_settings}. 
 }
    \label{tab:control_tasks}
\end{table*}

\paragraph{Baselines.}

Following the prior work \cite{liu2022composable}, we compare with FUDGE \cite{yang2021fudge} , Style
Transformer \cite{dai2019style}, and LatentOps \cite{liu2022composable} models to sequentially edit the source sentences as a baseline of
sequential editing. 
\ding{172} For FUDGE, it has a discriminator that takes in a prefix sequence and predicts whether the generated sequence
would meet the conditions.
\ding{173} Style transformer makes no assumption about the latent representation of source
sentence and takes the proven self-attention network. The Transformer at here serves as a basic module to train a
style transfer system.
\ding{174}  For LatentOps, following \citet{liu2022composable}, it permits plugging in attribute classifiers applied
on text latent vectors in the latent space and utilize
an ordinary differential equations sampler to draw latent
vector samples.

\subsection{Implementation Details}\label{sec:implementation_details}
% \sz{encoder decoder}
% \sz{flow}
Following the \citet{li2022diffusion} and \citet{ho2020denoising}, for language flow, we set the generation time steps to be 10 and the sequence length to be 64.
A U-Net \cite{ho2020denoising} backbone is utilized. 
% Parameters are shared across time, which is specified to the network using the Transformer sinusoidal position embedding \cite{vaswani2017attention}.
We use self-attention at the 32 feature map resolution \cite{wang2018non}. All models have two convolutional residual blocks per resolution level and self-attention blocks at the 16 resolution between the convolutional
blocks \cite{chen2018neural}. 
The generation time $t$ is specified by adding the Transformer sinusoidal position embedding into each residual block.
We train language flow using Adam optimizer and a
learning rate at $1 \times 10^{-5}$, dropout of 0.1, batch size of 64, and the total number of training iteration
is 20K for control tasks, and 30K for text editing tasks. For the details about the latent space structures, we include the details in Appendix~\ref{sec:app_exp}.

\section{Experiments}\label{sec:experiemental_results}
We evaluate the performance of our language rectified flow
and learning framework in this section. We bold the best result within each column block. The results of our method are obtained with three independent runs
to determine the variance. See Appendix~\ref{sec:app_exp} for full results with error bars.

\subsection{Control Tasks Results}\label{sec:control_tasks_section}
Table~\ref{tab:control_tasks} reports our results on three control-oriented text generation tasks. 
\ding{192}
The top block displays the performance of baselines and the {\ours} on the parts of speech and length, 
and the bottom block shows the results of infill task. We report the results with success metric (SR) and PPL. {\ours} shows consistent performance gains and better model generalization on both complex
controls task (Parts-of-speech) and
precise future planning tasks (Length) (\emph{e.g.}, $90.0 \rightarrow 94.2$ on success rate, $5.16 \rightarrow 4.65$ on fluency). 
\ding{193}
Our language rectified flow continually outperforms the baseline methods
for infilling. Our method achieves better performance in automatic evaluation.
These results suggest that {\ours} can solve many types of controllable generation tasks that
depend on generation order or lexical constraints (such as infilling) without specialized training.
\ding{194}
Through these results, it further confirms that {\ours} can work as an effective method to be incorporated into different-type models on the challenging fine-grained text generations.

\paragraph{Generation Efficiency.} 
We provide the running time of generating 50 examples for the parts of speech task. Experiments in this part
are performed on a single GPU during generation. The results of our method are tested with three independent runs and the average results are reported in Table \ref{tab:runningtime}. Our language flow with the domain transfer flow and straight through sampling is 26.7× faster than Diffusion LM and 16.7× faster than FUDGE.
It shows that {\ours} gives the best performance outperforming plug-and-play LM (FUDGE) and Diffusion LM (DLM), while keeping the lowest generation time.

% Thanks for the question. We add a compute cost section in the appendix (Appendix C), reporting the compute cost (clock time on a single GPU) to train and decode / control diffusion-lm. We also copy the content in this reply:

% Training diffusion LM (we trained it for 200k steps) takes around 8h on a single GPU. Training an autoregressive LM takes around 2h.

% Without control, sampling from a Diffusion-LM takes 63 seconds to obtain a batch of 50 samples. We experiment with some speed up by downsampling diffusion steps from 2000 to 200 at decoding time. The faster version takes 5.6s to obtain a batch of 50 samples. An autoregressive LM takes 0.7s to obtain a batch of 50 samples.

% Controllable generation of 50 samples takes around 80s for Diffusion-LM, 4800s for (unbatched) PPLM and 50s for FUDGE.

\begin{table}[h]
%  \footnotesize
\centering
\def\arraystretch{1.2}
\setlength\tabcolsep{18pt}
\resizebox{0.45\textwidth}{!}{\begin{tabular}{l|c|c|c}
 \toprule
 Model & FUDGE & DLM & Ours  \\ 
%  & {SQuAD1.1 + NQ}& { Other MRQA datasts}\\
 \hline
Time (s) & 50 & 80 & 3  \\ 
% Ours base  &51.3& 223M  & 11.2G \\
 \bottomrule
  \end{tabular}}
 \caption{Results of the generation time for each method. `s' represents the second.}
\label{tab:runningtime}
% \vspace{-15pt}
\end{table}

\vspace{-.4em}
\subsection{Text Editing}\label{sec:fact_ver_section}
% \sz{check Zhiting's paper is same setting that we can use for our baselines.}
% \sz{explain what the accuracy is in the table 4}
We further show the experimental results on the text editing in Table \ref{tab:text_edit_results}. We adopt several baselines from the existing literature. Following \citet{liu2022composable}, we compare our method with FUDGE \cite{yang2021fudge}, Style Transformer \cite{dai2019style} models, LatentOps \cite{liu2022composable}.
In Table \ref{tab:text_edit_results}: \ding{182}  We observe sizable gains over all baselines with a clear margin (from LatentOps' 24.1 to Ours 25.8).
\ding{183} {\ours} demonstrates the strong capability of controllable editing during the training and allowing the transport from the source distribution to the target distribution. Thus, it comes to the best performance in most of the settings.

\begin{table}[th]
 \footnotesize
\centering
\scalebox{0.8}{
 \begin{tabular}{l|cccc}
 \toprule
 Model 
 & BLEU $\uparrow$  &  Accuracy $\uparrow$   &  PPL $\downarrow$ \\
   \hline
 STrans \cite{dai2019style} & 25.6 & 0.89 & 41.4\\
 FUDGE \cite{yang2021fudge} & 17.2 & 0.36 & 38.9 \\
 LatentOps \cite{liu2022composable} & 24.1 & 0.93 & 26.1  \\
\hline
 {\bf Ours} & {\bf 25.8} & {\bf 0.95} & {\bf 24.2}\\
 \hline
 \hline
%   STrans \cite{dai2019style} & 24.5 & 41.0 & - & 57.9\\
 FUDGE \cite{yang2021fudge} & 35.3 & 0.25 & 50.2   \\
 LatentOps \cite{liu2022composable} & 28.7 & 0.77 & 44.8  \\
\hline
 {\bf Ours} & {\bf 39.9} & {\bf 0.86} & {\bf 40.1}\\
 \bottomrule
  \end{tabular}}
 \caption{
 Comparison to models on Yelp (top) and Amazon (bottom) dataset. Automatic evaluations (BLEU, accuracy, and PPL) are reported for each model. `STrans' represents style transformer. 
 }
    \label{tab:text_edit_results}
    \vspace{-10pt}
\end{table}

\section{Analysis}\label{sec:analysis_section}
\vspace{-.4em}
\paragraph{What is the influence of the constrained optimization vs. manually tuning coefficient?} 
Here we verify whether the constrained optimization in {\ours} is better than the manually tuning strategies.
With the designed trade-off, {\ours} targets optimizing between the flow and representation construction.  
Instead of automatically searching the trade-off between the the flow optimization and representation construction,
we manually set a constant $\lambda$ in Eqn \eqref{eq:our_pareto_2} as $0.1$, $1.0$, $2.0$. \ding{182} Table \ref{tab:ab_constrained_optimization} shows that 
the constrained optimization of our method brings clear benefits. \ding{183} We find that without CO, `$\minus$ CO' with different manually tuned $\lambda$ value shows an trade-off between the BLEU, ROUGE, and BERTScore across all $\lambda$ values, indicating that manually tuned $\lambda$ value can not bring the optimized flow and representation together.
% It is often beneficial to have the dynamically learned $\lambda$ from the constrained optimization.
It demonstrates the necessity and effectiveness of the constrained optimization for the switchable candidate set in {\ours} structure. We also study the impact of jointly training VAE and Unet vs. separately training them. More results are included in the Appendix~\ref{sec:app_exp}.

\begin{table}[th]
%  \footnotesize
\centering
\scalebox{0.65}{
 \begin{tabular}{l|c|c|c}
 \toprule
   & \multicolumn{3}{c}{Infill} \\ 
  \cline{2-4}
 & BLEU $\uparrow$ &  ROUGE-L $\uparrow$ & BERTScore $\uparrow$  \\
   \hline 
   DLM 
 & 7.1 & 28.3  & { 89.0 }  \\
{Ours} & {8.2} & {32.7} & {92.1} \\
- CO, $\lambda$ = 0.1 & 7.5 & 29.8  & 90.5 \\
 - CO, $\lambda$ = 1.0 & 6.5 & 28.8 & 88.3 \\
 - CO, $\lambda$ = 2.0 & 7.8 & 30.6 & 90.8 \\
 \bottomrule
  \end{tabular}}
 \caption{Comparison of different $\lambda$ values for the manually tuned trade-off between the flow and the representation vs. Ours. `CO' denotes constrained optimization.
 }
    \label{tab:ab_constrained_optimization}
    \vspace{-10pt}
\end{table}

\vspace{-.4em}
\paragraph{Ablations on the components of language rectified flow.}
Our language rectified flow focuses on latent diffusion. The latent space is constructed by VAE. Thereforce, the purpose of the latent flow is to learn the transport from the Gaussian distribution to the Gaussian distribution. Under this setting, UNet \cite{ho2020denoising} and transformer \cite{vaswani2017attention} are two eligible considerations. Following the diffusion language model hyperparameter and code base \cite{liu2022composable}, we obtain the below result (Table \ref{tab:unet_transform}) for the VAE + UNet (V+U) vs. VAE + transformer (V+T) on the infill task for our language rectified flow. It is clear that our LF can effectively utilizes VAE + UNet or VAE + transformer to achieve comparable performance. It indicates our method is insensitive to different learning structures.
This confirms our discussion in Section \ref{sec:method_section} that  {\ours} can serve as an efficient probability flow for language generations.

\begin{table}[h]
%  \footnotesize
\centering
 \resizebox{0.7\columnwidth}{!}{\begin{tabular}{l|c|c|c}
 \toprule
Model  & BLEU $\uparrow$ &  ROUGE-L $\uparrow$ & BERTScore $\uparrow$\\ 
  \cline{2-4}
   \hline
V+T & 7.9 & 32.9 & 92.0 \\  
V+U & 8.2 & {32.7} & {92.1}\\  
 \bottomrule
  \end{tabular}}
 \caption{Results of the impact of the latent flow with different model structures on the infill task.
 }
    \label{tab:unet_transform}
    \vspace{-10pt}
\end{table}

\vspace{-.4em}
\paragraph{Studies on the role of latent {\ours}.}
We conduct the ablation study to exam the role of latent LF in the latent space. 
With the designed flow transport, {\ours} targets the fine-grained text generation. 
We compare {\ours} with and w/o latent LF settings.
W/o latent LF here represents mapping the sentence in one domain to the latent space and directly mapping this latent code into a sentence in another domain without transporting by LF.
As shown in Table \ref{tab:influence_standard_dp}, the w/o latent LF strategy can not generate the high quality text while {\ours} yields better results with a clear margin.
It demonstrates the necessity and effectiveness of incorporating the flow with the domain transfer and faster sampling for text generation in {\ours} structure, and a possible reason is that it is too hard to optimize the discrete space.

\begin{table}[h]
%  \footnotesize
\centering
 \resizebox{0.8\columnwidth}{!}{\begin{tabular}{l|c|c|c}
 \toprule
Model  & BLEU $\uparrow$ &  ROUGE-L $\uparrow$ & BERTScore $\uparrow$\\ 
  \cline{2-4}
   \hline
W/o Latent Flow & 2.1 & 20.5 & 45.8\\  
Ours (w. latent Flow) & {8.2} & {32.7} & {92.1}\\  
 \bottomrule
  \end{tabular}}
 \caption{Ablation of the impact of without the latent flow on the infilling controllable text generation.
 }
    \label{tab:influence_standard_dp}
    \vspace{-15pt}
\end{table}

\begin{figure}[h]
    \centering
    \includegraphics[width=0.42\textwidth]{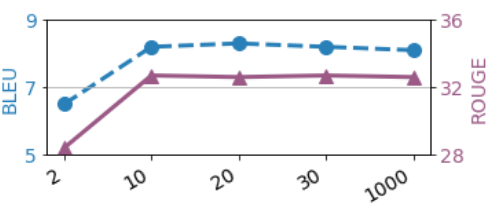}
    
    \caption{Results of the study on different generation steps for ours on the infill text generation. }
    \vspace{-15pt}
    \label{fig:ablation_steps}
\end{figure}

\paragraph{More evidence for faster simulations with different generation steps.}
As discussed in Section \ref{sec:eff_language_flow}, the propose language flow demonstrate the faster sampling with very few generation steps. In Section \ref{sec:implementation_details}, we consolidate the generation steps as ten by default. 
We select multiple generation steps and 
study {\ours}'s abilities in text generation with different schedule. 
We follow the same training settings in Section~\ref{sec:experiemental_settings} and present the results in Figure~\ref{fig:ablation_steps}. 
We notice that for our case the difference between different generation steps is small. 
Our method already converges well with just ten steps. The sample quality is further improved with additional update steps.

% A possible reason is that the number of class categories is small and therefore the choice of divergence does not have a large influence.
% In this part, we further study the performance by distilling the sampling steps into multiple steps similar to standard diffusion methods.
% We follow the same training settings in Section~\ref{exp:uncondition}.
% As we can see from Figure~\ref{fig:ablation}, 
% our method already converges well with just one step. The sample quality is further improved with additional update steps.

\vspace{-.4em}
\paragraph{Qualitative Analysis.}
% We show randomly sampled outputs of Diffusion-LM both for unconditional generation and for the 5
% control tasks. Table 8 shows the unconditional generation results. Table 9, Table 10, Table 12, and
% 18
% Table 3 show the qualitative samples from span control, POS control, semantic content control, and
% syntax tree control, respectively. Table 11 shows the results of length control.
We provide some generated examples in Table \ref{tab:qualititative_analysis} to raise a more
direct comparison. More examples are included in Appendix.
Consistent with the quantitative results, it is difficult for FUDGE to control all the
desired length while generate the the semantic informative sentence successfully. For example, the sentence's length of FUDGE is 13. 
Although Diffusion LM generate the informative sentence, it does not come with the perfect length as our target is 12 and the length of the generated sentence is 11. However, ours is inclined to
generate the informative and meaningful sentences with the perfect the length. In conclusion, there seem be a trade-off informative and the length. It
can be handled well by ours, but for the baselines, they either lose informative or the accurate length.

\begin{table}[h]
 \small
\centering
\def\arraystretch{1.2}
\setlength\tabcolsep{6pt}
 \resizebox{1.0\columnwidth}{!}{\begin{tabular}{l|c}
 \toprule
Model  &  Length: 12 \\
  % \cline{2-3}
   \hline
FUDGE & The Phoenix is an average Japanese restaurant that is in the City Centre. (13) \\  
Diffusion LM &  The Twenty Two serves Chinese food and is not family friendly. (11) \\
{\ours} & The sky glowed with a vibrant sunset, painting nature's masterpiece across the horizon. (12) \\
 \bottomrule
  \end{tabular}}
 \caption{Qualitative output of the length control tasks, where all the generated texts tried to match the target length exactly. We mark the length at the end of the
sentences.
 }
    \label{tab:qualititative_analysis}
    \vspace{-10pt}
\end{table}

\vspace{-.4em}
\section{Related Work}
\paragraph{Probabilistic Generative Models for Text}
Probabilistic models for text have shown promising performance improvements.
Previous methods \cite{mueller2017sequence,liu2020revision, fan2020bayesian, fan2021contextual} utilize the probabilistic variational auto encoders to encode the input sequence into the latent space and use networks for the jointly training.
\citet{mai2020plug} proposes to train an additional MLP in the latent space of an auto-encoder.
Because of the recent success of diffusion models, Diffusion LM \cite{li2022diffusion} and LDEBM \cite{yu2022latent} utilize the diffusion process in the latent space for text generation. For controllable text generation, \citet{yang2021fudge} learns an attribute predictor operating on a partial sequence. 
and \cite{xue2023parameter} uses parameter-efficient
tuning (e.g., prompting tuning and low-rank adaptation) to optimize control tokens for LLLM such as GPT and then fine-tune models for controllable generations. Our language rectified flow focuses on the domain transfer by utilizing the ordinary differential
equation (ODE) for the faster sampling, and we regard combining multiple constraints as a promising future direction. 
Our approach is built on the reformulation of the probabilistic flow. 
Similar to \citet{mai2020plug}, we build up a flow in the latent space constructed by an auto-encoder, and further transports the text input from the source distribution to the target distribution for the fine-grained and high-quality text generation.  

\vspace{-.4em}
\paragraph{Comparisons of Diffusion Flow and Flow-based Models.}
Existing flow-based text generation models are normalizing flow-based methods \cite{ma2019flowseq, ding2021flowprior, tang2021continuous}. Normalizing flows often require explicit, invertible transformations, often resulting in triangular or diagonal Jacobian matrices to ensure efficient computation of determinants. Thus, it needs careful design to ensure invertibility and tractable Jacobians making it hard to train. It requires designing an invertible layer in the neural network.
Our language flow is also an ODE / SDE probability flow \cite{song2021solving,lipman2022flow}. It has more flexibility to choose the model structures, such as the latest Unet or Transformer in our work. It is usually easy to train as we model the trajectory from the source distribution to the target distribution. Therefore our language flow has no constraints on both architecture and invertibility. In addition, LF and diffusion models can be viewed as members of the probability flow family.

\section{Conclusion}\label{sec:conclusion}
Our work demonstrates the benefits of introducing the language rectified flow. The proposed flow can learn the neural ordinary differential equation model to transport between the source distribution and the target distribution, providing the unified and effective solution to generative modeling and domain transfer. Our {\ours} produces the fine-grained and high quality controllable text with fast simulation.  
The proposed strategy 
shows noticeable gains in performance across controllable text generation (parts-of speech, length and infill) and text editing (Yelp and Amazon). We further conduct the detailed study with the LF in different settings, \emph{e.g.}, comparing between constrained optimization vs. manually tuning coefficient, providing more evidence for faster simulation with different generation steps, and verifying the impact of different components.
To summarize, the flow method is effective and general, with the potential to be incorporated into existing models for various NLP tasks.

% \section{Limitations}
% \label{sec:limitations}
% In real practices or real-life scenarios, the data is often biased. The gap between the training and testing data might be large and unexpected. Thus, incautious implementation or vague understanding of model output might lead to unanticipated false consequences. In addition, with computational consumption, environmentally sustainability and users friendly should be considered. 

\bibliography{naacl2021}

\begin{thebibliography}{59}
\expandafter\ifx\csname natexlab\endcsname\relax\def\natexlab#1{#1}\fi

\bibitem[{Anderson(1982)}]{anderson1982reverse}
Brian~DO Anderson. 1982.
\newblock Reverse-time diffusion equation models.
\newblock \emph{Stochastic Processes and their Applications}, 12(3):313--326.

\bibitem[{Bahdanau et~al.(2014)Bahdanau, Cho, and Bengio}]{bahdanau2014neural}
Dzmitry Bahdanau, Kyunghyun Cho, and Yoshua Bengio. 2014.
\newblock Neural machine translation by jointly learning to align and translate.
\newblock \emph{arXiv preprint arXiv:1409.0473}.

\bibitem[{Bhagavatula et~al.(2019)Bhagavatula, Bras, Malaviya, Sakaguchi, Holtzman, Rashkin, Downey, Yih, and Choi}]{bhagavatula2019abductive}
Chandra Bhagavatula, Ronan~Le Bras, Chaitanya Malaviya, Keisuke Sakaguchi, Ari Holtzman, Hannah Rashkin, Doug Downey, Scott Wen-tau Yih, and Yejin Choi. 2019.
\newblock Abductive commonsense reasoning.
\newblock \emph{arXiv preprint arXiv:1908.05739}.

\bibitem[{Bowman et~al.(2015)Bowman, Vilnis, Vinyals, Dai, Jozefowicz, and Bengio}]{bowman2015generating}
Samuel~R Bowman, Luke Vilnis, Oriol Vinyals, Andrew~M Dai, Rafal Jozefowicz, and Samy Bengio. 2015.
\newblock Generating sentences from a continuous space.
\newblock \emph{arXiv preprint arXiv:1511.06349}.

\bibitem[{Brown et~al.(2020)Brown, Mann, Ryder, Subbiah, Kaplan, Dhariwal, Neelakantan, Shyam, Sastry, Askell et~al.}]{brown2020language}
Tom Brown, Benjamin Mann, Nick Ryder, Melanie Subbiah, Jared~D Kaplan, Prafulla Dhariwal, Arvind Neelakantan, Pranav Shyam, Girish Sastry, Amanda Askell, et~al. 2020.
\newblock Language models are few-shot learners.
\newblock \emph{Advances in neural information processing systems}, 33:1877--1901.

\bibitem[{Chen et~al.(2018)Chen, Rubanova, Bettencourt, and Duvenaud}]{chen2018neural}
Ricky~TQ Chen, Yulia Rubanova, Jesse Bettencourt, and David~K Duvenaud. 2018.
\newblock Neural ordinary differential equations.
\newblock \emph{Advances in neural information processing systems}, 31.

\bibitem[{Cho et~al.(2014)Cho, Van~Merri{\"e}nboer, Gulcehre, Bahdanau, Bougares, Schwenk, and Bengio}]{cho2014learning}
Kyunghyun Cho, Bart Van~Merri{\"e}nboer, Caglar Gulcehre, Dzmitry Bahdanau, Fethi Bougares, Holger Schwenk, and Yoshua Bengio. 2014.
\newblock Learning phrase representations using rnn encoder-decoder for statistical machine translation.
\newblock \emph{arXiv preprint arXiv:1406.1078}.

\bibitem[{Chowdhery et~al.(2022)Chowdhery, Narang, Devlin, Bosma, Mishra, Roberts, Barham, Chung, Sutton, Gehrmann et~al.}]{chowdhery2022palm}
Aakanksha Chowdhery, Sharan Narang, Jacob Devlin, Maarten Bosma, Gaurav Mishra, Adam Roberts, Paul Barham, Hyung~Won Chung, Charles Sutton, Sebastian Gehrmann, et~al. 2022.
\newblock Palm: Scaling language modeling with pathways.
\newblock \emph{arXiv preprint arXiv:2204.02311}.

\bibitem[{Dai et~al.(2019)Dai, Liang, Qiu, and Huang}]{dai2019style}
Ning Dai, Jianze Liang, Xipeng Qiu, and Xuanjing Huang. 2019.
\newblock Style transformer: Unpaired text style transfer without disentangled latent representation.
\newblock \emph{arXiv preprint arXiv:1905.05621}.

\bibitem[{Dathathri et~al.(2019)Dathathri, Madotto, Lan, Hung, Frank, Molino, Yosinski, and Liu}]{dathathri2019plug}
Sumanth Dathathri, Andrea Madotto, Janice Lan, Jane Hung, Eric Frank, Piero Molino, Jason Yosinski, and Rosanne Liu. 2019.
\newblock Plug and play language models: A simple approach to controlled text generation.
\newblock \emph{arXiv preprint arXiv:1912.02164}.

\bibitem[{Dey and Salem(2017)}]{dey2017gate}
Rahul Dey and Fathi~M Salem. 2017.
\newblock Gate-variants of gated recurrent unit (gru) neural networks.
\newblock In \emph{2017 IEEE 60th international midwest symposium on circuits and systems (MWSCAS)}, pages 1597--1600. IEEE.

\bibitem[{Ding and Gimpel(2021)}]{ding2021flowprior}
Xiaoan Ding and Kevin Gimpel. 2021.
\newblock Flowprior: Learning expressive priors for latent variable sentence models.
\newblock In \emph{Proceedings of the 2021 Conference of the North American Chapter of the Association for Computational Linguistics: Human Language Technologies}, pages 3242--3258.

\bibitem[{Fan et~al.(2020)Fan, Zhang, Chen, and Zhou}]{fan2020bayesian}
Xinjie Fan, Shujian Zhang, Bo~Chen, and Mingyuan Zhou. 2020.
\newblock Bayesian attention modules.
\newblock \emph{Advances in Neural Information Processing Systems}, 33:16362--16376.

\bibitem[{Fan et~al.(2021)Fan, Zhang, Tanwisuth, Qian, and Zhou}]{fan2021contextual}
Xinjie Fan, Shujian Zhang, Korawat Tanwisuth, Xiaoning Qian, and Mingyuan Zhou. 2021.
\newblock Contextual dropout: An efficient sample-dependent dropout module.
\newblock \emph{arXiv preprint arXiv:2103.04181}.

\bibitem[{Goodfellow et~al.(2013)Goodfellow, Warde-Farley, Mirza, Courville, and Bengio}]{goodfellow2013maxout}
Ian Goodfellow, David Warde-Farley, Mehdi Mirza, Aaron Courville, and Yoshua Bengio. 2013.
\newblock Maxout networks.
\newblock In \emph{International conference on machine learning}, pages 1319--1327. PMLR.

\bibitem[{He and McAuley(2016)}]{he2016ups}
Ruining He and Julian McAuley. 2016.
\newblock Ups and downs: Modeling the visual evolution of fashion trends with one-class collaborative filtering.
\newblock In \emph{proceedings of the 25th international conference on world wide web}, pages 507--517.

\bibitem[{Ho et~al.(2020)Ho, Jain, and Abbeel}]{ho2020denoising}
Jonathan Ho, Ajay Jain, and Pieter Abbeel. 2020.
\newblock Denoising diffusion probabilistic models.
\newblock \emph{Advances in Neural Information Processing Systems}, 33:6840--6851.

\bibitem[{Karras et~al.(2022)Karras, Aittala, Aila, and Laine}]{karras2022elucidating}
Tero Karras, Miika Aittala, Timo Aila, and Samuli Laine. 2022.
\newblock Elucidating the design space of diffusion-based generative models.
\newblock \emph{arXiv preprint arXiv:2206.00364}.

\bibitem[{Kingma and Welling(2013)}]{kingma2013auto}
Diederik~P Kingma and Max Welling. 2013.
\newblock Auto-encoding variational bayes.
\newblock \emph{arXiv preprint arXiv:1312.6114}.

\bibitem[{Krause et~al.(2020)Krause, Gotmare, McCann, Keskar, Joty, Socher, and Rajani}]{krause2020gedi}
Ben Krause, Akhilesh~Deepak Gotmare, Bryan McCann, Nitish~Shirish Keskar, Shafiq Joty, Richard Socher, and Nazneen~Fatema Rajani. 2020.
\newblock Gedi: Generative discriminator guided sequence generation.
\newblock \emph{arXiv preprint arXiv:2009.06367}.

\bibitem[{Kumar et~al.(2021)Kumar, Malmi, Severyn, and Tsvetkov}]{kumar2021controlled}
Sachin Kumar, Eric Malmi, Aliaksei Severyn, and Yulia Tsvetkov. 2021.
\newblock Controlled text generation as continuous optimization with multiple constraints.
\newblock \emph{Advances in Neural Information Processing Systems}, 34:14542--14554.

\bibitem[{Li et~al.(2020)Li, Gao, Li, Peng, Li, Zhang, and Gao}]{li2020optimus}
Chunyuan Li, Xiang Gao, Yuan Li, Baolin Peng, Xiujun Li, Yizhe Zhang, and Jianfeng Gao. 2020.
\newblock Optimus: Organizing sentences via pre-trained modeling of a latent space.
\newblock \emph{arXiv preprint arXiv:2004.04092}.

\bibitem[{Li et~al.(2018)Li, Jia, He, and Liang}]{li2018delete}
Juncen Li, Robin Jia, He~He, and Percy Liang. 2018.
\newblock Delete, retrieve, generate: a simple approach to sentiment and style transfer.
\newblock \emph{arXiv preprint arXiv:1804.06437}.

\bibitem[{Li et~al.(2022)Li, Thickstun, Gulrajani, Liang, and Hashimoto}]{li2022diffusion}
Xiang~Lisa Li, John Thickstun, Ishaan Gulrajani, Percy Liang, and Tatsunori~B Hashimoto. 2022.
\newblock Diffusion-lm improves controllable text generation.
\newblock \emph{arXiv preprint arXiv:2205.14217}.

\bibitem[{Lin and Hovy(2003)}]{lin2003automatic}
Chin-Yew Lin and Eduard Hovy. 2003.
\newblock Automatic evaluation of summaries using n-gram co-occurrence statistics.
\newblock In \emph{Proceedings of the 2003 human language technology conference of the North American chapter of the association for computational linguistics}, pages 150--157.

\bibitem[{Lipman et~al.(2022)Lipman, Chen, Ben-Hamu, Nickel, and Le}]{lipman2022flow}
Yaron Lipman, Ricky~TQ Chen, Heli Ben-Hamu, Maximilian Nickel, and Matt Le. 2022.
\newblock Flow matching for generative modeling.
\newblock \emph{arXiv preprint arXiv:2210.02747}.

\bibitem[{Liu et~al.(2020)Liu, Fu, Zhang, Pal, and Lv}]{liu2020revision}
Dayiheng Liu, Jie Fu, Yidan Zhang, Chris Pal, and Jiancheng Lv. 2020.
\newblock Revision in continuous space: Unsupervised text style transfer without adversarial learning.
\newblock In \emph{Proceedings of the AAAI Conference on Artificial Intelligence}, volume~34, pages 8376--8383.

\bibitem[{Liu et~al.(2022{\natexlab{a}})Liu, Feng, Gao, Yang, Liang, Bao, He, Cui, Li, and Hu}]{liu2022composable}
Guangyi Liu, Zeyu Feng, Yuan Gao, Zichao Yang, Xiaodan Liang, Junwei Bao, Xiaodong He, Shuguang Cui, Zhen Li, and Zhiting Hu. 2022{\natexlab{a}}.
\newblock Composable text control operations in latent space with ordinary differential equations.
\newblock \emph{arXiv preprint arXiv:2208.00638}.

\bibitem[{Liu et~al.(2022{\natexlab{b}})Liu, Gong, and Liu}]{liu2022flow}
Xingchao Liu, Chengyue Gong, and Qiang Liu. 2022{\natexlab{b}}.
\newblock Flow straight and fast: Learning to generate and transfer data with rectified flow.
\newblock \emph{arXiv preprint arXiv:2209.03003}.

\bibitem[{Liu et~al.(2023{\natexlab{a}})Liu, Wu, Zhang, Gong, Ping, and Liu}]{liu2023flowgrad}
Xingchao Liu, Lemeng Wu, Shujian Zhang, Chengyue Gong, Wei Ping, and Qiang Liu. 2023{\natexlab{a}}.
\newblock Flowgrad: Controlling the output of generative odes with gradients.
\newblock In \emph{Proceedings of the IEEE/CVF Conference on Computer Vision and Pattern Recognition}, pages 24335--24344.

\bibitem[{Liu et~al.(2023{\natexlab{b}})Liu, Zhang, Ma, Peng et~al.}]{liu2023instaflow}
Xingchao Liu, Xiwen Zhang, Jianzhu Ma, Jian Peng, et~al. 2023{\natexlab{b}}.
\newblock Instaflow: One step is enough for high-quality diffusion-based text-to-image generation.
\newblock In \emph{The Twelfth International Conference on Learning Representations}.

\bibitem[{Ma et~al.(2019)Ma, Zhou, Li, Neubig, and Hovy}]{ma2019flowseq}
Xuezhe Ma, Chunting Zhou, Xian Li, Graham Neubig, and Eduard Hovy. 2019.
\newblock Flowseq: Non-autoregressive conditional sequence generation with generative flow.
\newblock \emph{arXiv preprint arXiv:1909.02480}.

\bibitem[{Mai et~al.(2020)Mai, Pappas, Montero, Smith, and Henderson}]{mai2020plug}
Florian Mai, Nikolaos Pappas, Ivan Montero, Noah~A Smith, and James Henderson. 2020.
\newblock Plug and play autoencoders for conditional text generation.
\newblock \emph{arXiv preprint arXiv:2010.02983}.

\bibitem[{Mireshghallah et~al.(2022)Mireshghallah, Goyal, and Berg-Kirkpatrick}]{mireshghallah2022mix}
Fatemehsadat Mireshghallah, Kartik Goyal, and Taylor Berg-Kirkpatrick. 2022.
\newblock Mix and match: Learning-free controllable text generation using energy language models.
\newblock \emph{arXiv preprint arXiv:2203.13299}.

\bibitem[{Mueller et~al.(2017)Mueller, Gifford, and Jaakkola}]{mueller2017sequence}
Jonas Mueller, David Gifford, and Tommi Jaakkola. 2017.
\newblock Sequence to better sequence: continuous revision of combinatorial structures.
\newblock In \emph{International Conference on Machine Learning}, pages 2536--2544. PMLR.

\bibitem[{Novikova et~al.(2017)Novikova, Du{\v{s}}ek, and Rieser}]{novikova2017e2e}
Jekaterina Novikova, Ond{\v{r}}ej Du{\v{s}}ek, and Verena Rieser. 2017.
\newblock The e2e dataset: New challenges for end-to-end generation.
\newblock \emph{arXiv preprint arXiv:1706.09254}.

\bibitem[{Papineni et~al.(2002)Papineni, Roukos, Ward, and Zhu}]{papineni2002bleu}
Kishore Papineni, Salim Roukos, Todd Ward, and Wei-Jing Zhu. 2002.
\newblock Bleu: a method for automatic evaluation of machine translation.
\newblock In \emph{Proceedings of the 40th annual meeting of the Association for Computational Linguistics}, pages 311--318.

\bibitem[{Pascanu et~al.(2013)Pascanu, Gulcehre, Cho, and Bengio}]{pascanu2013construct}
Razvan Pascanu, Caglar Gulcehre, Kyunghyun Cho, and Yoshua Bengio. 2013.
\newblock How to construct deep recurrent neural networks.
\newblock \emph{arXiv preprint arXiv:1312.6026}.

\bibitem[{Qin et~al.(2020)Qin, Shwartz, West, Bhagavatula, Hwang, Bras, Bosselut, and Choi}]{qin2020back}
Lianhui Qin, Vered Shwartz, Peter West, Chandra Bhagavatula, Jena Hwang, Ronan~Le Bras, Antoine Bosselut, and Yejin Choi. 2020.
\newblock Back to the future: Unsupervised backprop-based decoding for counterfactual and abductive commonsense reasoning.
\newblock \emph{arXiv preprint arXiv:2010.05906}.

\bibitem[{Qin et~al.(2022)Qin, Welleck, Khashabi, and Choi}]{qin2022cold}
Lianhui Qin, Sean Welleck, Daniel Khashabi, and Yejin Choi. 2022.
\newblock Cold decoding: Energy-based constrained text generation with langevin dynamics.
\newblock \emph{arXiv preprint arXiv:2202.11705}.

\bibitem[{Radford et~al.(2019)Radford, Wu, Child, Luan, Amodei, Sutskever et~al.}]{radford2019language}
Alec Radford, Jeffrey Wu, Rewon Child, David Luan, Dario Amodei, Ilya Sutskever, et~al. 2019.
\newblock Language models are unsupervised multitask learners.
\newblock \emph{OpenAI blog}, 1(8):9.

\bibitem[{Rombach et~al.(2022)Rombach, Blattmann, Lorenz, Esser, and Ommer}]{rombach2022high}
Robin Rombach, Andreas Blattmann, Dominik Lorenz, Patrick Esser, and Bj{\"o}rn Ommer. 2022.
\newblock High-resolution image synthesis with latent diffusion models.
\newblock In \emph{Proceedings of the IEEE/CVF Conference on Computer Vision and Pattern Recognition}, pages 10684--10695.

\bibitem[{Ronneberger et~al.(2015)Ronneberger, Fischer, and Brox}]{ronneberger2015u}
Olaf Ronneberger, Philipp Fischer, and Thomas Brox. 2015.
\newblock U-net: Convolutional networks for biomedical image segmentation.
\newblock In \emph{Medical Image Computing and Computer-Assisted Intervention--MICCAI 2015: 18th International Conference, Munich, Germany, October 5-9, 2015, Proceedings, Part III 18}, pages 234--241. Springer.

\bibitem[{Shen et~al.(2017)Shen, Lei, Barzilay, and Jaakkola}]{shen2017style}
Tianxiao Shen, Tao Lei, Regina Barzilay, and Tommi Jaakkola. 2017.
\newblock Style transfer from non-parallel text by cross-alignment.
\newblock \emph{Advances in neural information processing systems}, 30.

\bibitem[{Song et~al.(2021)Song, Shen, Xing, and Ermon}]{song2021solving}
Yang Song, Liyue Shen, Lei Xing, and Stefano Ermon. 2021.
\newblock Solving inverse problems in medical imaging with score-based generative models.
\newblock \emph{arXiv preprint arXiv:2111.08005}.

\bibitem[{Song et~al.(2020)Song, Sohl-Dickstein, Kingma, Kumar, Ermon, and Poole}]{song2020score}
Yang Song, Jascha Sohl-Dickstein, Diederik~P Kingma, Abhishek Kumar, Stefano Ermon, and Ben Poole. 2020.
\newblock Score-based generative modeling through stochastic differential equations.
\newblock \emph{arXiv preprint arXiv:2011.13456}.

\bibitem[{Tang et~al.(2021)Tang, Zhang, Kim, and Bansal}]{tang2021continuous}
Zineng Tang, Shiyue Zhang, Hyounghun Kim, and Mohit Bansal. 2021.
\newblock Continuous language generative flow.
\newblock In \emph{Proceedings of the 59th Annual Meeting of the Association for Computational Linguistics and the 11th International Joint Conference on Natural Language Processing (Volume 1: Long Papers)}, pages 4609--4622.

\bibitem[{Vaswani et~al.(2017)Vaswani, Shazeer, Parmar, Uszkoreit, Jones, Gomez, Kaiser, and Polosukhin}]{vaswani2017attention}
Ashish Vaswani, Noam Shazeer, Niki Parmar, Jakob Uszkoreit, Llion Jones, Aidan~N Gomez, {\L}ukasz Kaiser, and Illia Polosukhin. 2017.
\newblock Attention is all you need.
\newblock \emph{Advances in neural information processing systems}, 30.

\bibitem[{Wang et~al.(2018)Wang, Girshick, Gupta, and He}]{wang2018non}
Xiaolong Wang, Ross Girshick, Abhinav Gupta, and Kaiming He. 2018.
\newblock Non-local neural networks.
\newblock In \emph{Proceedings of the IEEE conference on computer vision and pattern recognition}, pages 7794--7803.

\bibitem[{Xue et~al.(2023)Xue, Wang, and Ji}]{xue2023parameter}
Tianci Xue, Ziqi Wang, and Heng Ji. 2023.
\newblock Parameter-efficient tuning helps language model alignment.
\newblock \emph{arXiv preprint arXiv:2310.00819}.

\bibitem[{Yang and Klein(2021)}]{yang2021fudge}
Kevin Yang and Dan Klein. 2021.
\newblock Fudge: Controlled text generation with future discriminators.
\newblock \emph{arXiv preprint arXiv:2104.05218}.

\bibitem[{Yu et~al.(2022)Yu, Xie, Ma, Jia, Pang, Gao, Zhu, Zhu, and Wu}]{yu2022latent}
Peiyu Yu, Sirui Xie, Xiaojian Ma, Baoxiong Jia, Bo~Pang, Ruigi Gao, Yixin Zhu, Song-Chun Zhu, and Ying~Nian Wu. 2022.
\newblock Latent diffusion energy-based model for interpretable text modeling.
\newblock \emph{arXiv preprint arXiv:2206.05895}.

\bibitem[{Zhang et~al.(2021{\natexlab{a}})Zhang, Gong, and Choi}]{zhang2021knowing}
Shujian Zhang, Chengyue Gong, and Eunsol Choi. 2021{\natexlab{a}}.
\newblock Knowing more about questions can help: Improving calibration in question answering.
\newblock \emph{arXiv preprint arXiv:2106.01494}.

\bibitem[{Zhang et~al.(2021{\natexlab{b}})Zhang, Gong, and Choi}]{zhang2021learning}
Shujian Zhang, Chengyue Gong, and Eunsol Choi. 2021{\natexlab{b}}.
\newblock Learning with different amounts of annotation: From zero to many labels.
\newblock \emph{arXiv preprint arXiv:2109.04408}.

\bibitem[{Zhang et~al.(2022{\natexlab{a}})Zhang, Gong, and Liu}]{zhang2022passage}
Shujian Zhang, Chengyue Gong, and Xingchao Liu. 2022{\natexlab{a}}.
\newblock Passage-mask: A learnable regularization strategy for retriever-reader models.
\newblock \emph{arXiv preprint arXiv:2211.00915}.

\bibitem[{Zhang et~al.(2022{\natexlab{b}})Zhang, Gong, Liu, He, Chen, and Zhou}]{zhang2022allsh}
Shujian Zhang, Chengyue Gong, Xingchao Liu, Pengcheng He, Weizhu Chen, and Mingyuan Zhou. 2022{\natexlab{b}}.
\newblock Allsh: Active learning guided by local sensitivity and hardness.
\newblock \emph{arXiv preprint arXiv:2205.04980}.

\bibitem[{Zhang et~al.(2023)Zhang, Gong, Wu, Liu, and Zhou}]{zhang2023automl}
Shujian Zhang, Chengyue Gong, Lemeng Wu, Xingchao Liu, and Mingyuan Zhou. 2023.
\newblock Automl-gpt: Automatic machine learning with gpt.
\newblock \emph{arXiv preprint arXiv:2305.02499}.

\bibitem[{Zhang et~al.(2022{\natexlab{c}})Zhang, Roller, Goyal, Artetxe, Chen, Chen, Dewan, Diab, Li, Lin et~al.}]{zhang2022opt}
Susan Zhang, Stephen Roller, Naman Goyal, Mikel Artetxe, Moya Chen, Shuohui Chen, Christopher Dewan, Mona Diab, Xian Li, Xi~Victoria Lin, et~al. 2022{\natexlab{c}}.
\newblock Opt: Open pre-trained transformer language models.
\newblock \emph{arXiv preprint arXiv:2205.01068}.

\bibitem[{Zhang et~al.(2019)Zhang, Kishore, Wu, Weinberger, and Artzi}]{zhang2019bertscore}
Tianyi Zhang, Varsha Kishore, Felix Wu, Kilian~Q Weinberger, and Yoav Artzi. 2019.
\newblock Bertscore: Evaluating text generation with bert.
\newblock \emph{arXiv preprint arXiv:1904.09675}.

\end{thebibliography}
\bibliographystyle{acl_natbib}
\clearpage

\clearpage
% \onecolumn
\appendix
\section{Experimental details}\label{sec:app_exp}
\subsection{Full Results With Error Bar }\label{sec:appendix_fullresults}
We report the full results of our method with the error bar for control tasks and text editing tasks in Table \ref{tab:appendix_control_tasks} and  \ref{tab:appendix_text_edit_results}, respectively.

\begin{table}[th]
%  \footnotesize
\centering
\scalebox{0.70}{
 \begin{tabular}{l|cccc}
 \toprule
\multirow{2}{*}{Model}  & \multicolumn{2}{c}{Parts of Speech} & \multicolumn{2}{c}{Length}\\ 
  \cline{2-5}
 & SR $\uparrow$ &  PPL $\downarrow$ & SR $\uparrow$  &  PPL $\downarrow$\\
   \hline
%  PPLM \cite{dathathri2019plug} & - & - & - & 57.9\\
 FUDGE 
%  \cite{yang2021fudge} 
 & 27.0 & 7.96 & 46.9 & 3.11 \\
 FT 
%  \cite{radford2019language}  
 & 89.5 & 4.72 & 98.1 & 3.84 \\
 DLM 
%  \cite{li2022diffusion} 
 & 90.0 & 5.16  & {99.9} & 2.16  \\
\hline
  {Ours} & {94.2$\pm$0.4} & {4.65$\pm$0.1} & {99.3$\pm$0.2} &{1.74$\pm$0.2} \\
  \hline
  \hline
  \multicolumn{2}{c|}{~} & \multicolumn{3}{|c}{Infill} \\ 
  \cline{3-5}
  \multicolumn{2}{c|}{~} & BLEU $\uparrow$ &  ROUGE-L $\uparrow$ & BERTScore $\uparrow$\\
   \hline 
    \multicolumn{2}{c|}{DELOREAN}
%  \cite{yang2021fudge} 
 & 1.6 & 19.1 & 41.7 \\
 \multicolumn{2}{c|}{COLD} 
%  \cite{radford2019language}  
 & 1.8 & 19.5 &42.7 \\
 \multicolumn{2}{c|}{DLM} 
%  \cite{li2022diffusion} 
 & 7.1 & 28.3  & { 89.0 } \\
 \hline
 \multicolumn{2}{c|}{Ours} & {8.2$\pm$0.3} & {32.7$\pm$0.4} & {92.1$\pm$0.2}\\
 \bottomrule
  \end{tabular}}
 \caption{Comparison to models on Parts of Speech, Length, and Infill. `SR' and ` PPL' refer to the success metric and the fluency of the generated text, in Section \ref{sec:control_experiemental_settings}. 
 }
    \label{tab:appendix_control_tasks}
    % \vspace{-10pt}
\end{table}

\begin{table}[th]
 \footnotesize
\centering
\scalebox{0.8}{
 \begin{tabular}{l|cccc}
 \toprule
 Model 
 & BLEU $\uparrow$  &  Accuracy $\uparrow$   &  PPL $\downarrow$ \\
   \hline
 STrans \cite{dai2019style} & 25.6 & 0.89 & 41.4\\
 FUDGE \cite{yang2021fudge} & 17.2 & 0.36 & 38.9 \\
 LatentOps \cite{liu2022composable} & 24.1 & 0.93 & 26.1  \\
\hline
 {Ours} & {25.8$\pm$0.2} & {0.95$\pm$0.1} & {24.2$\pm$0.3}\\
 \hline
 \hline
%   STrans \cite{dai2019style} & 24.5 & 41.0 & - & 57.9\\
 FUDGE \cite{yang2021fudge} & 35.3 & 0.25 & 50.2   \\
 LatentOps \cite{liu2022composable} & 28.7 & 0.77 & 44.8  \\
\hline
 {Ours} & {39.9$\pm$0.5} & {0.86$\pm$0.05} & {40.1$\pm$0.3}\\
 \bottomrule
  \end{tabular}}
 \caption{
 Comparison to models on Yelp (top) and Amazon (bottom) dataset. Automatic evaluations (BLEU, accuracy, and PPL) are reported for each model. `STrans' represents style transformer. 
 }
    \label{tab:appendix_text_edit_results}
    % \vspace{-10pt}
\end{table}

\subsection{Experimental Datasets}
For parts-of-speech and length in control tasks, we rely on the E2E dataset \cite{novikova2017e2e}. The infill task is based on Abductive NLG dataset \cite{bhagavatula2019abductive}.
For text editing tasks, we utilize the Yelp review \cite{shen2017style} and Amazon comment corpus \cite{he2016ups}.

{\bf E2E dataset} \cite{novikova2017e2e} was assembled using the CrowdFlower platform and underwent quality control according to \citet{novikova2017e2e}. This dataset contains information about restaurants and comprises over 50k combinations of dialogue-act-based MRs with an average of 8.1 references each. The dataset is divided into training, validation, and testing sets (at a 76.5-8.5-15 ratio), maintaining a similar distribution of MR and reference text lengths, and ensuring that MRs in different sets are unique. Each MR features 3-8 attributes (slots), such as name, food, or area, along with their corresponding values. In line with \citet{novikova2017e2e}, the E2E data was collected using images as stimuli, which were found to produce significantly more natural, informative, and well-phrased human references than textual MRs. {\bf Abductive NLG (aNLG)} is a task focused on generating natural language explanations for given observations. It is built on the ART dataset \cite{bhagavatula2019abductive}, which contains more than 20k commonsense narrative contexts and 200k explanations. The training set comprises both plausible and implausible hypotheses gathered through crowdsourcing. In contrast, the development and test sets include hypotheses chosen using the Adversarial Filtering algorithm.
{\bf Yelp review} dataset is sourced from the 2018 Yelp Challenge, which focuses on local businesses such as restaurants and bars, treating them as items. To maintain data quality, the same 10-core setting is employed. The preprocessed Yelp review data utilized in this study is provided by \citet{li2018delete}.
The {\bf Amazon dataset} shares similarities with Yelp. Each example consists of a sentence from an Amazon product review and is labeled as expressing either positive or negative sentiment \cite{he2016ups}.

\subsection{Experimental Settings}
We use a U-Net \cite{ho2020denoising} backbone similar to an unmasked PixelCNN++ \cite{ronneberger2015u}.
We set gradient clipping to 1.0.
For control tasks, the fine-tune GPT-2 \cite{radford2019language, zhang2022allsh} on (control, text) pair. We report the sampling (with temperature 1.0) of the fine-tuned models denoted as FT. For FUDGE \cite{yang2021fudge}, it incorporates a discriminator that examines a prefix sequence and anticipates whether the generated sequence will comply with the conditions. It can steer text generation by adjusting the probabilities of the pretrained language model based on the discriminator's output. Following \citet{li2022diffusion}, we adopt FUDGE's architecture, training a discriminator for each attribute. A three-layer LSTM followed by a Linear layer is served as the discriminator.
For the latent space encoder and decoder, we employ an RNN encoder–decoder setup, as suggested by \citet{cho2014learning, bahdanau2014neural}. The RNN Encoder–Decoder used in our experiment contains 1K hidden units, featuring the proposed gates in both the encoder and decoder.

The encoder of the our latent construction comprises forward and backward RNNs, each with 1K hidden units. The decoder also contains 1K hidden units. In both instances, we utilize a multilayer network with a single maxout hidden layer, as per \citet{goodfellow2013maxout}, to calculate the conditional probability of each target word, following the method outlined by \citet{pascanu2013construct}. We have implemented the GRU as the activation function, as suggested by \citet{dey2017gate}. Attention is integrated into the decoder, allowing it to determine the sections of the source sentence that should be focused on, in accordance with \citet{bahdanau2014neural}. We maintain the same optimizer and learning rate in this setup. The language rectified flow experiments are carried out in an end-to-end fashion, as described by \citet{rombach2022high}.

For the latent space encoder and decoder, we train an RNN encoder–decoder \cite{cho2014learning, bahdanau2014neural}. 
The RNN Encoder–Decoder used in the experiment had 1K hidden units with the proposed
gates at the encoder and at the decoder. 
The encoder of the
RNN consists of forward and backward recurrent neural networks (RNN) each having 1K
hidden units. Its decoder has 1K hidden units. In both cases, we use a multilayer network with a
single maxout \cite{goodfellow2013maxout} hidden layer to compute the conditional probability of each
target word \cite{pascanu2013construct}. The GRU \cite{dey2017gate} is implemented as the activation function. The attention is incorporated in the decoder.
The decoder decides parts of the source
sentence to pay attention to \cite{bahdanau2014neural}. We use the same optimizer and learning rate here. The language flow experiments are conducted in an end-to-end manner \cite{zhang2021knowing, zhang2021learning, rombach2022high}.
More detailed experimental settings are included in Appendix \ref{sec:app_exp}.

\subsection{Jointly Train vs. Separately Train.}
Training the language rectified flow and VAE separately requires more training time. In Table \ref{tab:training_seperately}, We obtain the below result for the infill task. It is clear that our LF effectively utilizes constrained optimization to achieve slightly better performance while still maintaining the lower training cost (training iterations). From experimental results, we can successfully train them together, and we hypothesize that our VAE is easier to train under these settings.

\begin{table}[h]
%  \footnotesize
\centering
 \resizebox{0.9\columnwidth}{!}{\begin{tabular}{l|c|c|c|c}
 \toprule
Model  & BLEU $\uparrow$ &  ROUGE-L $\uparrow$ & BERTScore $\uparrow$ & Number of Training iterations $\downarrow$\\ 
  \cline{2-5}
   \hline
Separately & 7.9 & 32.5 & 92.2 & VAE: 10K + LF 20K \\  
Ours & 8.2 & {32.7} & {92.1} & 20K\\  
 \bottomrule
  \end{tabular}}
 \caption{Reported results of Training LF and VAE.
 }
    \label{tab:training_seperately}
    \vspace{-10pt}
\end{table}

\subsection{More Qualitative Examples}
We provide some generated examples in Table \ref{tab:appendix_qualititative_analysis} to raise a more
direct comparison. Aligned with prior quantitative findings, there appears to be a trade-off between informativeness and sentence length. Our model manages this well, but the baseline models tend to sacrifice either informativeness or precise length.

\begin{table}[h]
 \small
\centering
 \resizebox{1.0\columnwidth}{!}{\begin{tabular}{l|c}
 \toprule
Model &  Length: 6 \\
  % \cline{2-3}
   \hline
FUDGE & Climate change affects global weather patterns. (6) \\  
FUDGE & Recent virtual reality technology is advancing rapidly. (7) \\  
FUDGE & Blockchain is revolutionizing financial systems. (5)\\  
\hline
Diffusion LM &  Artificial intelligence transforms modern healthcare. (5) \\
Diffusion LM & Plant-based diets gain popularity worldwide. (6) \\
Diffusion LM &  Remote work reshapes traditional office culture. (6) \\
\hline
{\ours} & Technology enhances learning experiences in schools. (6)\\
{\ours} & The global economy navigates unprecedented challenges. (6)\\
{\ours} & Sustainability remains crucial in modern architecture. (6) \\
 \bottomrule
  \end{tabular}}
 \caption{More qualitative output of the length control tasks, where all the generated texts try to match the target length exactly. We mark the length at the end of the sentences. 
 }
    \label{tab:appendix_qualititative_analysis}
    % \vspace{-15pt}
\end{table}

\end{document}